\documentclass{article} 

\date{}
\usepackage{microtype}
\usepackage{graphicx}
\usepackage{booktabs} 
\usepackage{hyperref}
\usepackage{url}
\usepackage{amsmath}
\usepackage{algorithm}
\usepackage{algorithmic}
\usepackage{amssymb}
\usepackage{tikz}
\usepackage{appendix}
\usepackage{subcaption}
\usepackage{color}
\usepackage{xcolor}
\usepackage{indentfirst}
\usepackage{authblk}
\newtheorem{theorem}{Theorem}[section]
\newtheorem{lemma}{Lemma}[section]

\newtheorem{definition}{Definition}[section]

\newtheorem{assumption}{Assumption}[section]
\title{Positively Scale-Invariant Flatness of ReLU Neural Networks}
\author[1,2]{Mingyang Yi}
\author[3]{Qi Meng}
\author[3]{Wei Chen}
\author[1,2]{Zhi-ming Ma}
\author[3]{Tie-Yan Liu}
\affil[1]{Academy of Mathematics and Systems Science, Chinese Academy of Sciences}
\affil[2]{University of Chinese Academy of Sciences}
\affil[3]{Microsoft Research}
\affil[ ]{yimingyang17@mails.ucas.edu.cn, mazm@amt.ac.cn, \{meq,wche, tie-yan.liu\}@microsoft.com}

\begin{document}
\maketitle

	\begin{abstract}
		It was empirically confirmed by Keskar et al.\cite{SharpMinima} that flatter minima generalize better. However, for the popular ReLU network, sharp minimum can also generalize well \cite{SharpMinimacan}. The conclusion demonstrates that the existing definitions of flatness fail to account for the complex geometry of ReLU neural networks because they can't cover the Positively Scale-Invariant (PSI) property of ReLU network. In this paper, we formalize the PSI causes problem of existing definitions of flatness and propose a new description of flatness - \emph{PSI-flatness}. 
		PSI-flatness is defined on the values of basis paths \cite{GSGD} instead of weights. Values of basis paths have been shown to be the PSI-variables and can sufficiently represent the ReLU neural networks which ensure the PSI property of PSI-flatness. Then we study the relation between PSI-flatness and generalization theoretically and empirically. 
		First, we formulate a generalization bound based on PSI-flatness which shows generalization error decreasing with the ratio between the largest basis path value and the smallest basis path value. That is to say, the minimum with balanced values of basis paths will more likely to be flatter and generalize better. Finally. we visualize the PSI-flatness of loss surface around two learned models which indicates the minimum with smaller PSI-flatness can indeed generalize better. 

	\end{abstract}
	\section{Introduction}
	DNN (i.e. DNN) with rectifier linear units (ReLU) activation has achieved huge success in quite a lot fields of machine learning, such as computer vision \cite{VGG, lenet,he2016deep,huang2017densely}, natural language processing \cite{NLP1, NLP2,vaswani2017attention}, information system \cite{cheng2016wide, wang2017deep}, etc. It is surprising that DNN can generalize well to unseen data despite its overwhelming capacity. Explaining this phenomenon is a challenging topic, because the classical uniform bound based on Rademacher complexity or VC dimension for analyzing generalization error seems to be too loose in practical \cite{zhang2016understanding}. Many works try to seek new measures that related to the generalization for DNN \cite{SharpMinima,neyshabur2017exploring,hardt2015train}.   
	
	\par
	It is arguably believed that a flat minimum of loss function found by stochastic gradient based methods results in good generalization ability \cite{SharpMinima,loss-surface2}. 
	A minimum is flat if there is a wide region around it with roughly the same error. However, the existing definitions of flatness \cite{SharpMinima, three_factors, Wu} are argued by Dinh et al. \cite{SharpMinimacan} that for ReLU neural network, there is no specific link between these definitions of flatness and generalization error by constructing counter examples based on the positive homogeneity of ReLU function. The counter examples reveal that minima can still generalize well even with large flatness value\footnote{When the flatness measure has large value, it means the loss surface is sharp.}. Hence, existing definitions of flatness fail to account for the complex geometry of ReLU NNs, because a well-defined measure of flatness should at least cover the positive homogeneity of ReLU network. 
	\par
	It's well known that ReLU NN is positively scale-invariant (PSI) \cite{PathSGD} which is a more general property than positive homogeneity. More specific, for a ReLU network, PSI means the ingoing weights of one hidden node multiplied by a positive constant $c$ and the outgoing weights of the hidden node divided by $c$ at the same time while the output of network keeps unchanged under such transformation. 
	Recent works \cite{PathSGD,GSGD,GSGD-norm} have been concerned to cope with such property using paths or basis paths of ReLU NN. 
	
	We analyze the exiting flatness and show that if the definitions of flatness of ReLU NN dissatisfy PSI, there are two networks with same generalization error but different measures of flatness. Thus such definitions can be an inappropriate measure. We claim that well-defined flatness should cover the PSI property of ReLU network, 
	 and a function of PSI-variables will naturally be PSI. Hence, it's suitable to measure flatness of RelU network via some PSI-variables. 
	
	In fact, values of basis paths have been proved to be PSI-variables \cite{GSGD}. We leverage them to define a PSI-flatness for ReLU network. Specifically, a path is the connection between an input node and an output node of a ReLU network. The value of a path is defined as the product of weights along the path. And basis paths are the maximal linearly independent subset of all paths (the details will be introduced in Section \ref{sec: back}). We propose three definitions of PSI-flatness accord with three generally accepted flatness in weight space. We uniformly call them PSI-flatness which reveal the local variation of the loss surface of model represented by values of basis paths instead of weights. Compared with the original definitions \cite{SharpMinima,three_factors,BehnamNeyshabur}, PSI-flatness captures the PSI property of ReLU networks. Hence, we claim that PSI-flatness is a suitable definition measure of flatness.
	\par
	We further confirm the relationship between PSI-flatness and generalization error. 
	Theoretically, we leverage a PAC Bayes generalization error bound to bridge PSI-flatness and generalization ability. We show that minima with small ratio between the largest basis path value and the smallest basis path value correspond with small PSI-flatness therefore better generalization. 
	Empirically, we visualize the loss surface represented by basis path values following similar visualization strategy in weight space proposed by Goodfellow et al.\cite{Visualize}.
	We compare the loss surface around a well-generalized point and a poor-generalized point. The results show that the point generalize well indeed locates in a flatter valley. The experimental results confirm our conclusion that minima with smaller PSI-flatness have better generalization ability.

	\subsection{Related Work}
	Flatness is a popular topic to understand generalization both empirically and theoretically. Experimental results \cite{SharpMinima, loss-surface2} illustrate that SGD with large batch size tends to produce ''sharp" minimum that generalize worse. Based on this observation, quite a lot measures are proposed to describe flatness \cite{SharpMinima, three_factors, BehnamNeyshabur, Bayes}. Unfortunately, the existing definitions of flatness are questioned by Dinh et al.\cite{SharpMinimacan}. They construct counterexamples to illustrate that the definitions of flatness are ill-posed. Hence they suggest that an appropriate measure should consider reparameterization.
	\par
	There are some works focus on the path value based reparameterization of ReLU NN. Neyshabur et al.\cite{PathSGD} introduce the positively scaling transformation and firstly propose the path parameterized method. 
	Because the path values have overlaps, Meng et al.\cite{GSGD} show that a subset of path values called basis path values, are sufficient to represent ReLU networks. They claim the space composed by basis path values is a PSI space. 
	\par
	Existing references related to the two topics are mutually independent. None of these papers consider using a reparameterization method to describe flatness appropriately. In this paper, we study how PSI property of ReLU will influence the definition of geometrical measures, and leverage values of basis paths to propose the proper definition of flatness. In addition, we analyze the properties of PSI-flatness theoretically and empirically.
	
	\section{Background}\label{sec: back}
	
	\subsection{Existing Definitions of Flatness}
	Given the loss function $l(\cdot)$ and a minimum $w$, there are quite a lot general accepted measures to describe flatness around a minimum. We focus in three definitions. They respectively are $\varepsilon$-weight flatness \cite{SharpMinima}; trace-weight flatness \cite{three_factors} and expected-weight flatness \cite{neyshabur2017exploring}. 
	In this paper, if there is no extra illustration, the loss $l(\cdot)$ means empirically loss.  
	\begin{definition}[Weight flatness]\label{def1}
		Let $B_{2}(w,\varepsilon)$ be an Euclidean ball centered on a minimum $w$ with radius $\varepsilon$. Then, for a non-negative valued loss function $l(\cdot)$, the $\varepsilon$-weight flatness will be defined as
		\begin{equation}
		F_{\varepsilon\emph{-weight}}(w)=\frac{\mathop{\max}_{w^{\prime}\in B_{2}(w,\varepsilon)}\left(l(w^{\prime})-l(w)\right)}{1+l(w)}.
		\label{sharpness}
		\end{equation}
		Meanwhile, we can define trace-weight flatness $F_{\emph{Hessian-weight}}(H)$ as $\text{tr}(H)$ ($H$ is the Hessian of $w$). And expected-weight flatness $F_{\emph{expected-weight}}(w,u)$ on point $w$ is $|E_{u}[l(w+u)]-l(w)|$, where $u$ is a random vector.
	\end{definition}
	From the definitions of weight flatness, we see that minima with small weight flatness are flat minima.

	\subsection{Basis Paths of ReLU NN}\label{sec2.1}
	\begin{figure}[t]\centering
		\includegraphics[width=0.25\columnwidth]{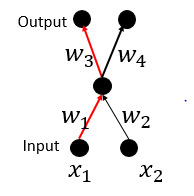}
		\caption{This is a simple ReLU network with one hidden node. Suppose path values are {$v_{1}(w)=w_1w_3, v_{2}(w)=w_1w_4, v_{3}(w)=w_2w_3, v_{4}(w)=w_2w_4$}, we can see the inner-dependency between them, i.e., {$v_{4}(w)=\frac{v_{2}(w)\cdot v_{3}(w)}{v_{1}(w)}$}. }
		\label{fig:network}
	\end{figure}
	The outputs of ReLU neural NN with $L$ layer can be calculated as $f_w(x):=w_L\cdot\sigma_{relu}(w_{L-1}\cdot\sigma_{relu}(\cdots \sigma_{relu}(w_1x)))$, where $\sigma_{relu}(\cdot)$ is the entry-wised non-linear ReLU activation function. The ReLU NN can be regarded as a directed graph. We use $p$ to denote a path starting from an input node and pointing to an output node by crossing one hidden node at each layer. $\mathcal{P}_{k,i}$ denotes the set of paths starting from the $i$-th input and pointing to the $k$-th output. Then the ($k$-th) output can be represented as 
	\begin{equation}
	f^{k}_{v}(x) = \sum_{i}\sum_{p\in\mathcal{P}_{k,i}}v(p)\cdot a(w,x) \cdot x_{i},
	\label{G-output}
	\end{equation}  
	where $v(p)=\prod_{l=1}^L w_l^{i_l}$ is the product of weights that the path $p$ passes by. And $a(w,x)=\prod_{l=1}^{L-1}I(o_l^{i_l}>0)$ where $o_l^{i_l}$ is the $i_l$-th output at layer $l$ that path $p$ passed by. We call $v(p)$ path value and $a(w,x)$ activation status. 
	\par
\begin{figure}\centering
		\includegraphics[width=0.3\columnwidth]{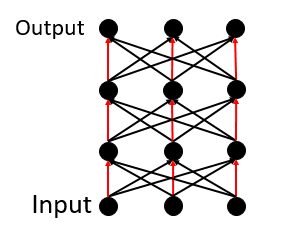}
		\caption{This is a MLP network with equal width. The red lines of the network are skeleton weights. The path contains at most one red line is basis path.}
		\label{fig:network2}
	\end{figure}

	However, path values are overlapped with each other (See Figure \ref{fig:network}). Meng et al.\cite{GSGD} disentangle the relation of paths and give a more compact representation of ReLU network. They regard a path as a vector with same dimension to the number of all the parameters of network, where each element represents whether the weight $w_j$ is passed by the path or not. If $w_j$ is passed by path $p$, the $j$-th component of vector $p$ denoted by $p_j$ is $1$; otherwise, $p_j=0$. 
	All the path vectors of a ReLU network compose a matrix which is denoted as $A$. The basis path is defined as follows.
	\begin{definition}[Basis Path]\cite{GSGD}
		A subset $\mathcal{P}_{0}$ of total paths $\mathcal{P}$ is basis path set if the path vectors in it compose a maximal linearly independent group of matrix $A$. 
	\end{definition}
	Here, we simply describe the \emph{skeleton method} to identify the basis paths for multilayer perceptron, which is designed in Meng et al.\cite{GSGD}. It first identify the skeleton weights. For MLP with equally width, the skeleton weights are the diagonal elements in the weight matrix at each layer. Then the paths that contain at most one skeleton weights are basis paths. Figure \ref{fig:network2} shows an example. The red lines are skeleton weights. Based on equation \eqref{G-output} and Theorem 3.6 in Meng et al.\cite{GSGD}, the loss function $l(w)$ can be defined in the basis value space which perform as $l(v)$ as long as the free-skeleton weights keep the signal unchanged. The free-skeleton weights are actually the read line not in the first layer of Figure \ref{fig:network2}, our later conclusion will build under such circumstance.  
	
	Basis paths constructed using this method can be classified into two types. One contains only skeleton weights and the other type contains one non-skeleton weights. When the values of free skeleton weights are fixed, there is a bijection between weights and basis paths. We use $p_{j}$ to denote the basis path which contains one non-skeleton weight $w_{j}$.
	We use $p_{i}$ to denote the basis paths which contains only skeleton weights. Among all the weights $p_i$ contains,  $w_{i}$ is the one at first layer. 
	We will use the mapping between $p_i$ and $w_i$, and the mapping between $p_j$ and $w_j$ in Section \ref{sec:the} and Section \ref{sec:exp}. 
	
	\section{Analysis of Existing Flatness}
	
	One important property of ReLU activation function is the positively scale-invariant property: $c\cdot\sigma_{\text{relu}}(z)=c\cdot \max\{z,0\}=\sigma_{\text{relu}}(c\cdot z)$. Dinh et al.\cite{SharpMinimacan} point out that given a ReLU NN, the outputs of network are invariant under the transformation: $T_{c}: (w_1,w_2)\rightarrow (c\cdot w_1,\frac{1}{c}\cdot w_2), c>0$, where $(w_{1},w_{2})$ is the weight matrix of two adjacent layers. It makes the infinite set of weights $\{T_c(w_1,w_2), \forall c| \textit{Given}\ (w_1,w_2)\}$ unidentifiable. Using $T_c$, Dinh et al.\cite{SharpMinimacan} construct minima generalize well but correspond with infinite weight flatness  which reveals that these definitions are uninformative to generalization error.  
	
	
	We consider a more general transformation - positively scaling transformation. Meanwhile, we claim that weight flatness is ill-posed under such more extensive circumstance. First, we give a detailed description of positively scale transformation. For a hidden neuron $O_{i}$ of a NN, the $\phi_{c,O_{i}}(w)$ transforms the input weights $w_{O_{i}}$ to $c\cdot w_{O_{i}}$ and output weights $w_{O_{i}}^{\prime}$ as $\frac{1}{c}\cdot w_{O_{i}}^{\prime}$, where $c>0$. For all hidden neurons $\{O_{1},\cdots,O_{H}\}$ of ReLU NN, a positively scaling transformation $\phi_{c}(\cdot):\mathcal{W}\rightarrow\mathcal{W}$ is defined as: 
	\begin{equation}
	\phi_{c}(w) = \phi_{c_{1},O_{1}}\circ \cdots\phi_{c_{H},O_{H}}(w),
	\end{equation}
	for a vector $c=(c_{1},\cdots,c_{H})$, where $\circ$ means function composition.
	
	Obviously, the family of transformations $T_{c}$ defined in Definition 5 of Dinh et al.\cite{SharpMinimacan} is a subclass of positively scaling transformations. So the following results obtained under positively scaling transformations are stronger and naturally satisfied for $T_c$.

	
	We define the invariant variables of positively scaling transformations as follows.
	\begin{definition}\label{iv}
		If a function $g(w): \mathcal{W}\rightarrow \mathbb{R}$ satisfy $g(w)=g(\phi_c(w)), \forall c$, we say $g$ is an invariant variable of positively scaling transformations, abbreviated as PSI-variable.
	\end{definition}
	Because PSI property of ReLU network, the loss functions $l(w)$ is PSI-variables. Thus, the properties related with loss functions (such as generalization ability) need to consider the PSI property of network. The next theorem shows that if a measure of ReLU NN is not PSI, then it is unsuitable to be used to analyze the generalization error.
	\begin{theorem}\label{thm3.1}
		Given an ReLU NN model $f_w$. If the measure $h(w)$ is not invariant  to all the positively scaling transformations, there exists another model $f_{w'}$ satisfy $h(w)\neq h(w')$ and $f_w(x)=f_{w'}(x)$ for any input $x$.
	\end{theorem}
	\par
	\textbf{Proof}.
	Suppose $h(w)\neq h(\phi_{c}(w))$ for a specific $c$. By choosing $w^{\prime}=\phi_{c}(w)$ for $c$, we can easily verify that $f_{w}(x)=f_{w^{\prime}}(x)$ for any input $x$ by PSI property of ReLU network. Then we get the conclusion.$\hfill$ $\blacksquare$
	\par
	Combining Definition \ref{iv} of the flatness in Section \ref{sec2.1} and Theorem \ref{thm3.1}, we have the following theorem which reveals the weight flatness is not suitable because they are not PSI-variables.
	\begin{theorem}\label{prop3.1}
		The three weight flatness defined in Definition \ref{def1} are not PSI-variables.
	\end{theorem}
	\par
	\textbf{Proof}.
	First, we prove the non-PSI property of $F_{\varepsilon\text{-weight}}(w)$.
	Suppose that $O_{1,s},\cdots,O_{h,s}$ are the hidden nodes at layer-$s$. For fixed weight vector $w$ with non-zero elements, there exist a positively scaling transformation $\phi_c(w)$ with $c=(1,\cdots,1,c_{1,s},\cdots,c_{h,s},1,\cdots,1)$ to make $w'=\phi_c(w)$. We use $w_{i,s}$ and $w_{i,s+1}$ to denote the incoming and outgoing weight vector of node $O_{i,s}$ respectively. Making $c_{i,s}=\frac{\epsilon}{\sqrt{h}\|w_{i,s}\|}$, we have $(w_1,\cdots,w_{s-1},0,w_{s+1}',\cdots,w_L)\in B(w',\epsilon)$. So $\max_{B(w',\epsilon)}l(w')$ is at least as high as constant function, which makes the value of $F_{\varepsilon\text{-weight}}(w)$ relatively large. 
	\par
	For $F_{\text{Hessian-weight}}(H)=\text{tr}(H)$, we choose $c$ as $(\alpha,1,\cdots,1)$. Without loss of generality, we assume the weight parameter $w$ is represented as $(w_{O_{1}},w_{O_{1}}^{\prime},\cdots)$. Then, we have 
	\begin{equation}
	\begin{aligned}
	\nabla^{2}l(\phi_{c}(w)) = 
	\left[\begin{matrix}
	\alpha^{-1} I_{n_{1}} & & \\
	& \alpha I_{n_{1}} & \\
	& & I_{n_{2}}
	\end{matrix}\right]\nabla^{2} l(w) \left[\begin{matrix}
	\alpha^{-1} I_{n_{1}} & & \\
	& \alpha I_{n_{1}} & \\
	& & I_{n_{2}}
	\end{matrix}\right],
	\end{aligned}
	\end{equation}
	where $n_{1}$ is the dimension of vector $w_{O_{1}}$ and $n_{2}$ is the dimension of parameters un-related to neuron $O_{1}$. Denoting $\nabla^{2} l(w)$ as 
	\begin{equation}
	\left[\begin{matrix}
	A_{1} & * & *\\
	* & A_{2} & *\\
	* & * & A_{3}
	\end{matrix}\right],
	\end{equation}
	where $A_{1},A_{2}$ have same dimension with $I_{n_{1}}$; and $A_{3}$ has same dimension with $I_{n_{2}}$. We have 
	\begin{equation}
	\text{tr}\left(\nabla^{2}l(\phi_{c}(w))\right) = \text{tr}(\alpha^{-2}A_{1}+\alpha^{2}A_{2}+A_{3}).
	\end{equation}
	Since $\nabla^{2}l(w)$ is the Hessian of a minimum point $w$, we have $\text{tr}(A_{1})>0$. Let $\alpha$ goes to zero, then $\text{tr}(\nabla^{2}l(\phi_{c}(w)))\rightarrow \infty$. But $\text{tr}(\nabla^{2} l(w)) = \text{tr}(A_{1}+A_{2}+A_{3})$ is a finite number.
	\par
	Finally, for expected flatness $|E_{u}[l(w+u)]-l(w)|$, as long as $u$ take negative number. We can construct a $w^{\prime}=\phi_{c}(w)$ like in the proof of $\varepsilon$-flatness which ensure 
	\[(w_1,\cdots,w_{s-1},0,w_{s+1}',\cdots,w_L)\] locates in the line between $u+\phi_{c}(w)$ and $\phi_{c}(w)$ as long as $u$ has negative value. According to the continuity of $l(\cdot)$, we can achieve a relatively large expected flatness 
	$\hfill\blacksquare$       
	\par
	
	Theorem \ref{prop3.1} illustrates that the ill-posed problem of weight flatness is not only rely on the positive homogeneity property of ReLU function, but also caused by a more general PSI property of ReLU function. Hence, just like Dinh et al.\cite{SharpMinimacan} discussed: an appropriate measurement of flatness needs considering defining flatness in some reparameterized space. More specifically, we point that ReLU network should consider \emph{reparameterized PSI-variable} to describe flatness.

	\par
	\section{PSI-flatness}
	We have already discussed that PSI property of ReLU network causes the ill-posed problem of existing definitions of flatness. Hence, in this section, we aim to seek a positively scale-invariant measure to reflect the flatness of loss surface of ReLU NN. 
	\par
	The following theorem shows that a measure of ReLU model is PSI if it is defined on the invariant variables to positively scaling transformation. 
	\begin{theorem}
		Given an ReLU NN model $f_w$ and a group of variables $g_i: \mathcal{W}\rightarrow\mathbb{R}$, $i=1,\cdots,z$ that are invariant to all the positively scaling transformations. If the measure $h$ is a function of these variables, i.e., $h(g_1,\cdots,g_z)$, then $h$ has positively scale-invariant property.	\label{thm:equivalent} 
	\end{theorem}
	\par
	\vbox{}
	\noindent\textbf{Proof}.
	If $g_i(w), i=1,\cdots,z$ are PSI-variables, they satisfy that $g_i(w)=g_i(\phi_c(w))$ for any positively scaling transformation $\phi_c(\cdot)$. So we have 
	\[h(g_1(w),\cdots,g_z(w))=h(g_1(\phi_c(w)),\cdots,g_z(\phi_c(w))).\] That means $h$ has positively scale-invariant property.
	\par
	According to Theorem \ref{thm:equivalent}, defining a measure on the PSI-variables is a way to find the suitable definition of flatness. There are lots of variables satisfy PSI. However, the variables we need should be sufficient to represent the ReLU NN. That is to say, the loss function can be totally calculated by these variables. For example, the set of constant functions $g(w)=c$ is a counter example, which is composed by PSI-variables but not sufficient to represent the loss function. 
	\par
	Fortunately, Meng et al.\cite{GSGD} have proven that values of basis paths are PSI-variables and sufficient to represent the ReLU NN. 
	Thus, values of basis paths described in Section \ref{sec2.1} are exactly what we need. It has been proved that the number of basis paths is $m-H$, where $m$ is the dimension of weight vector $w$ and $H$ is the number of hidden nodes. Let $v=(v_1,\cdots,v_{m-H})$ to denote the basis path value vector. We correct the definition of weight flatness in Definition \ref{def1} by defining them on values of basis paths. 
	\begin{definition}[PSI-flatness]
		Representing ReLU NN by values of basis paths as $f_{v}$. Let $B_{2}(v,\varepsilon)$ be an Euclidean ball centered on a minimum $v$ with radius $\varepsilon$. Then, for a non-negative valued loss function $l(f_{v};x,y)$, the PSI-$\varepsilon$-flatness is defined as
		\begin{equation}
		F_{\emph{PSI-}\epsilon}(v)=\frac{\mathop{\max}_{v^{\prime}\in B_{2}(v,\varepsilon)}\left(l(v^{\prime})-l(v)\right)}{1+l(v)}.
		\label{g-sharpness}
		\end{equation}
		Meanwhile, PSI-trace flatness $F_{\emph{PSI-Hessian}}(H)$ is $\text{tr}(H)$, where $H$ is the Hessian matrix of $v$. And the PSI-expected flatness $F_{\emph{PSI-expected}}(v, u)$ of $v$ is $|E_{u}[l(v+u)]-l(v)|$, where $l(\cdot)$ is loss function and $u$ is a random vector.
	\end{definition} 

	We collectively call the three definitions of flatness as PSI-flatness. From the definition, we see minima with small PSI-faltness are flat minima. We name them with the prefix PSI because the space composed by values of basis paths is named as PSI-space.
	\par
	According to Theorem \ref{thm:equivalent} and PSI property of basis paths value, three types of PSI-flatness are all positively scale-invariant. Therefore, they capture the PSI geometrical property of the loss surface for ReLU networks. It's suitable to leverage them to study the generalization ability of ReLU NN. 
	\par
	Till now, we give an positive answer to Dinh et al.\cite{SharpMinimacan}: we find a definition of flatness for ReLU NN that has PSI property.

	\section{Generalization Error Bound Based on PSI-flatness}\label{sec:the}
	Though we give an appropriate measure of flatness, the relationship between PSI-flatness and generalization is uncleared up till now. A perspective of weight flatness connecting generalization error is revealed by Neyshabur et al.\cite{BehnamNeyshabur}, which relies on PAC Bayes theory \cite{PAC1,PAC2}. Motivated by their study, the relationship between PSI-flatness and generalization can be derived.
	
	In fact, given a prior distribution $P$ over the hypothesis space that is independent of the training data, with probability at least $1-\delta$, we have
	\begin{equation}
	\begin{aligned}
	\mathbb{E}_{u}[l(v+u)] - \hat{l}(v) \leq F_{\text{PSI-expected}}(v,u)+ 4\sqrt{\frac{1}{n}\left(KL(v+u||P)+\log{\frac{2n}{\delta}}\right)}.
	\end{aligned}
	\label{eq:g-generalization_bound}
	\end{equation}
	where $l(\cdot)$ and $\hat{l}(\cdot)$ are respectively expected loss and empirical loss, $n$ is the number of training data, basis path value $v$ is the minimum of $\hat{l}(\cdot)$ learned from the training data. We would like to give some explanations about this generalization bound. The generalization error is described by perturbed generalization error $\mathbb{E}_{u}[l(v+u)] - \hat{l}(v)$ rather the exact generalization error $l(v) - \hat{l}(v)$. A small perturbation variable $u$ will make the perturbed generalization error close to the real generalization error which involves a small $\|u\|$. However, if the norm of $u$ tends to zero, the KL term in equation \eqref{eq:g-generalization_bound} will goes to infinity and make the bound vacuous. Hence, it causes a trade off when we chose the $\|u\|$ under perspective of equation \eqref{eq:g-generalization_bound}. Even though, the result relatively gives a quantitatively description of generalization error.
	
	Next we analyze the relationship of three PSI-flatness which ensure we can analyze the three PSI-flatness at the same time.  
	\par
	On one hand, by Taylor's expansion we can easily derive that PSI-$\varepsilon$ flatness is larger than $\varepsilon^{2}\frac{\text{tr}(H(v))}{d}$,
	where $H(v)$ is Hessian matrix of $v$. On the other hand, if the perturbation $u$ in PSI-expected flatness satisfies $\|u\|\leq\varepsilon$, PSI-$\varepsilon$ flatness is larger than PSI-expected flatness. To summary, PSI-$\varepsilon$ flatness is the strongest measure to describe "flatness". Hence, combining equation \eqref{eq:g-generalization_bound} and an upper bound to PSI-$\varepsilon$ flatness can describe generalization error. 
	
	
	
	We use $v$ to denote the vector composed by values of basis paths. If $v$ is a minimum, the denominator of PSI-$\varepsilon$ flatness defined in equation \eqref{g-sharpness} will be close to zero. Hence, we use 
	\begin{equation}
	\mathop{\max}_{v^{\prime}\in B_{2}(v,\varepsilon)}\hat{l}(v^{\prime})-\hat{l}(v)
	\label{eq:g-flatness-subsitute}
	\end{equation}
	to approximate the PSI-$\varepsilon$ flatness, where $\hat{l}(\cdot)$ is empirically loss. Next, we will analyze the upper bound of $\mathop{\max}_{v^{\prime}\in B_{2}(v,\varepsilon)}\hat{l}(v^{\prime})-\hat{l}(v)$. We need the following assumptions to derive our result.
	\par
	\begin{assumption}
		The $L_2$ norm of the input of every layer can be upper bounded by a constant $C$. 
		\label{A1}
	\end{assumption}
	
	\par
	\begin{assumption}
		The loss function $l(f_v(x),y)$ is Lipschitz continuous to the output vector:
		\begin{align}\label{Lipsch}
		|l(f_v(x),y)-l(f_{v'}(x),y)|\leq  C_L\|f_v(x)-f_{v'}(x)\|
		\end{align}
		\label{A2}
	\end{assumption} 
	\begin{theorem}
		Under Assumption \ref{A1} and \ref{A2}, for a $L$-hidden layer ReLU NN with $d_{0}$ dimensional input and width $d_1$ in each hidden layer, we have
		\begin{equation}
		\begin{aligned}
		\mathop{\max}_{v^{\prime}\in B_{2}(v,\varepsilon)}\hat{l}(v^{\prime})-\hat{l}(v)\leq 2C\cdot C_{L}\cdot d_0\cdot d_{1}^{L-1}\cdot  d_{L}\cdot \left(1+2^{\frac{3L+4}{2}}(L-1)d_{1}d_L\psi(v)\right)\cdot\varepsilon,
		\end{aligned}
		\end{equation}
		Here $\psi(v)=\sup_{s\in[L]}\varepsilon^{s-1}\frac{\sup_{i\in[m-H]}|v_{i}|^{2L+1-s}}{\inf_{i\in[m-H]}|v_{i}|^{2L}}$ \footnote{$\sup_{s\in[L]}$ denotes the maximum among $s=1$, $s=2,\cdots,s=L$} and $v_{i}$ is the $i$-th basis path value. And $\varepsilon$ is smaller than a constant decided by path value and input data.
		\label{thm:bound}
	\end{theorem}
	Actually, we can get exactly scale of $\varepsilon$. It can be referred to the detailed proof of the theorem in Appendix. 
	\par
	From the theorem, we can analyze how the structure of the NN and the values of basis paths will influence the $F_{\textit{PSI-}\epsilon}$. As the size (both the width and depth) of the network becomes large, the value of $F_{\textit{PSI-}\epsilon}$  becomes large which means the loss surface will become sharp. 
	The ratio of the largest basis path value and the smallest basis path value $\frac{\sup_{i\in[m-H]}|v_{i}|}{\inf_{i\in[m-H]}|v_{i}|}$ which reflects the variability of basis path values also influences $F_{\textit{PSI-}\epsilon}$.  More specifically, a minimum of NN equipped with balanced basis path values is a flatter minimum which leads to better generalization.
	\par
	We will give a proof sketch of Theorem \ref{thm:bound}. We need a lemma to reveal how to use the values of basis paths to calculate the values of non-basis paths.  The following lemma is motivated by Proposition 1 in \cite{GSGD-norm}. We formulate it as follows.
	\begin{lemma}
		For a NN with $L$ layers and the number of  basis paths is $m-H$, the value of every non-basis path can be represented as:
		\begin{equation}
		\frac{v_{1}^{a_{1}}\cdots v_{m_{1}}^{a_{m_{1}}}}{v_{m_{1}+1}^{a_{m_{1}+1}}\cdots v_{m-H}^{a_{m-H}}},
		\label{eq:non-basispath}
		\end{equation}
		where $v_{1},\cdots,v_{m-H}$ are values of basis paths. $a_{1},\cdots,a_{m-H}$ in equation \eqref{eq:non-basispath} are non-negative integers satisfy $a_{1}+\cdots+a_{m_{1}}=L+1$ and $a_{m_{1}+1}+\cdots+a_{m-H}=L$. 
		\label{lem:repr}
	\end{lemma}
	\par 
	
	The lemma gives an explicit expression of non-basis path values. It helps to estimate the difference between $\hat{l}(v^{\prime})$ and $\hat{l}(v)$ in Theorem \ref{thm:bound}.
	\par
	Here we give a sketch of the proof for Theorem \ref{thm:bound}, the detailed proof can be found in supplement material.
	\par
	\vbox{}\noindent\textbf{Sketch of Proof for Theorem \ref{thm:bound}:}
	The main purpose is to bound $\hat{l}(v^{\prime})-\hat{l}(v)$ when $\|v^{\prime}-v\|\leq\varepsilon$. 
	Suppose that $v^{\prime}$ achieve the maximum in Theorem \ref{thm:bound} satisfies $v{'}-v=\delta$ with $\|\delta\|\leq\epsilon$. Then, we divide the proof into three steps.
	\par
	\noindent\textbf{Step 1}: By Assumption \ref{A2}, we have 
	\begin{equation}
	\begin{aligned}
	|l(f_{v}(x),y)-l(f_{v^{\prime}}(x),y)|\leq C_{L}\|f_{v}(x) - f_{v^{\prime}}(x)\|.
	\label{eq:l_continuous}
	\end{aligned}
	\end{equation}
	For all input data $x$, if the output of each neuron in the network keep same signal after perturbation, then the activation status keeps unchanged. Under such condition, by the equation \eqref{G-output}, we have 
	\begin{equation}
	\begin{aligned}
	\|f_{v^{\prime}}(x) - f_v(x)\|&=\left|\sum_{k=1}^{K}\left(\sum_{i}\sum_{p\in\mathcal{P}_{k,i}}(v_p^{\prime}-v_p)\cdot a_p(v,x) \cdot x_{i}\right)^{2}\right|^{\frac{1}{2}}\\
	&\leq \left|\sum_{k=1}^{K}n_{p}^{2}\left(\sum_{i}\sum_{p\in\mathcal{P}_{k,i}}(v^{\prime}_{p}-v_{p})^{2}a_{p}(v;x)\cdot x_{i}^{2}\right)\right|^{\frac{1}{2}},
	\label{eq:outputbound}
	\end{aligned}
	\end{equation}
	where $n_{p}$ is the number of total paths.
	According to the definition of empirically loss, we can bound the difference of output as well as $v_{p}^{\prime}-v_{p}$. Next, we control the scale of $\varepsilon$ to ensure the activation status unchanged for all input $x$ and give a bound to variable $v_{p}^{\prime}-v_{p}$.
	\par
	\vbox{}
	\noindent\textbf{Step 2}: The activation status of a path is decided by the outputs of the hidden neurons it crossed. For all input data $x$, if the output of each neuron in the network keep same signal after perturbation, then the activation status keeps unchanged. By controlling the scale of $\varepsilon$, we can achieve this.
	\par
	Let $D_{i}$ represent the indicator vector with the same dimension to the number of hidden nodes at layer $l$. For example, $D_{1}=(1,1,0,\cdots,0)$ represents the first two hidden neurons of layer 1 are activated while others are not. We use $D^{\prime}_{i}$ to be the indicator vector for the NN after perturbation.  Next, we analyze the condition of $\delta$ to make $\|D^{\prime}_{i}-D_{i}\|^{2} = 0$ for any $i$ and input $x$. 
	We have
	\begin{equation}
	\begin{aligned}
	\|D_{1}-D_{1}^{\prime}\|^{2}
	&=\sum_{i=1}^{d_1}\left|\mathbb{I}\left(\sum_{j=1}^{d_0}(v_{j}^{i}+\delta_{j}^{i})\cdot x_{j}\geq 0\right)-\mathbb{I}\left(\sum_{j=1}^{d_0}v_{j}^{i}\cdot x_{j}\geq 0\right)\right|\\
	&= \sum_{i=1}^{d_1}\mathbb{I}\left(\sum_{j=1}^{d_0}(v_{j}^{i}+\delta_{j}^{i})\cdot x_{j}\cdot\sum_{j=1}^{d_0}v_{j}^{i}\cdot x_{j}\leq 0\right),
	\end{aligned}
	\label{eq:signal}
	\end{equation}
	where $v_{j}^{i}$ is the value of basis path connecting the $i$-th hidden node in layer 1 and the $j$-th input node, $\delta_{j}^{i}$ is the perturbation to its corresponding path value. We can derive the upper bound for $\delta$ which forces the $\|D_{1}^{\prime}-D_{1}\|$ in equation \eqref{eq:signal} equals to zero. Similarly, we can give an analysis to $\delta$ which makes $\|D_{i}^{\prime}-D_{i}\|=0$. After that, we obtain the condition of unchanged activation status after perturbation.
	\par
    \vbox{}\noindent\textbf{Step 3}: Since we want to bound the difference of the path values for the NN before and after the perturbation. It's easily to verify the bound to $v_{p}^{\prime}-v_{p}$ for basis path. Based on Lemma \ref{lem:repr}, for any non-basis path $p$, the difference of path value is
	\begin{equation}
	\left|\frac{\prod_{i=1}^{L+1}(v_{p_{i}}+\delta_{p_{i}})}{\prod_{i=L+2}^{2L+1}(v_{p_{i}}+\delta_{p_{i}})} - \frac{\prod_{i=1}^{L+1}v_{p_{i}}}{\prod_{i=L+2}^{2L+1}v_{p_{i}}}\right|,
	\label{eq:non-basis-path-perturbation}
	\end{equation}
	where $v_{p_{1}},\cdots v_{p_{2L+1}}$ are corresponding basis paths value in Lemma \ref{lem:repr}; and $\delta_{p_{1}},\cdots \delta_{p_{2L+1}}$ are perturbations. We can derive an upper bound to \eqref{eq:non-basis-path-perturbation} based on basis path value and $\delta$. $\hfill\blacksquare$
	
	\section{Visualizing Loss Landscapes}\label{sec:exp}
	In this section, we will visualize the loss landscape represented by basis path values to study the relationship between generalization and PSI-flatness. 
	To study the relationship between flatness and generalization, previous work \cite{loss-surface2} has visualized the loss landscapes around minima in weight space.
	However, the results can's be applied directly to PSI-flatness, because loss function is defined in a totally different space. In order to further confirm the claim that: Minimum with smaller PSI-flatness generalize better, we visualize the loss landscape represented by basis path values around the minimum found by stochastic gradient descent (SGD). 
	
	\par
	First, we present the visualization method for PSI-flatness. We see the training loss $\hat{l}(\cdot)$ of a ReLU NN can be represented by basis path value vector $v$ as 
	\begin{equation}
	\hat{l}(v)=\frac{1}{n}\sum_{i=1}^{n}l(v,x_{i},y_{i}),
	\end{equation} 
	where $(x_{i},y_{i})$ is training data. Since $v$ is usually a high dimensional variable, visualization is possible only for one or two dimension. Here we discuss two dimensional contour plots of loss landscape.
	
	A general approach of visualizing two dimensional loss landscape is "Random Direction" which was used in \cite{Visualize}. We generalize the method to PSI space. For a model $f_{v^{*}}$, we choose two random direction vectors with dimension $m-H$ (number of basis paths), $\xi$ and $\eta$, which are independently sampled from Gaussian distribution. 
	Then we plot the graph of function
	\begin{equation}
	f(t_{1},t_{2}) = l(v^{*} + t_{1}\xi + t_{2}\eta),
	\label{eq:visualize}
	\end{equation}
	with coordinate system $t_{1}$ and $t_{2}$. Obviously, visualizing $f(t_1,t_2)$ is to visualize the loss function after adding perturbation to basis path values. Intuitively, a flatter landscape is more robust to perturbations, so the plot of $f(t_1,t_2)$ should be flatter.
	\par
	\begin{figure*}[t]\centering
		\begin{subfigure}[b]{0.23\textwidth}
			\includegraphics[width=\textwidth]{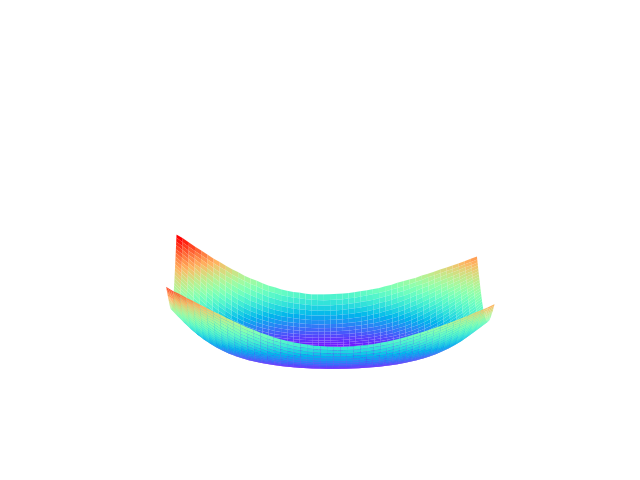}
			\caption{$128$; $87.78\%$}
			\label{fig:surface_1}
		\end{subfigure}
		\begin{subfigure}[b]{0.23\textwidth}
			\includegraphics[width=\textwidth]{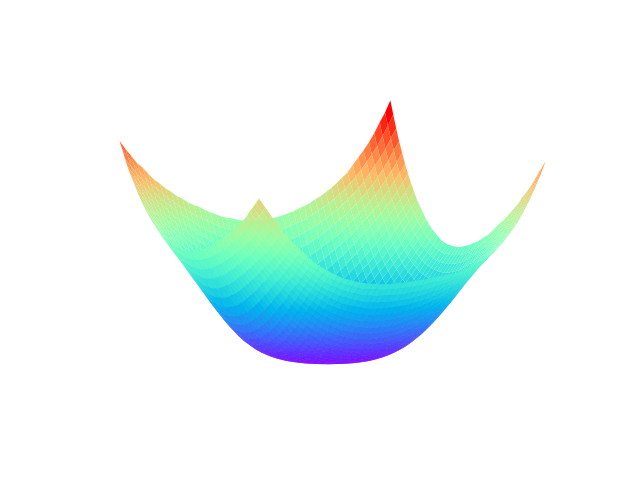}
			\caption{$2048$; $86.3\%$}
			\label{fig:surface_2}
		\end{subfigure}
		\begin{subfigure}[b]{0.23\textwidth}
			\includegraphics[width=\textwidth]{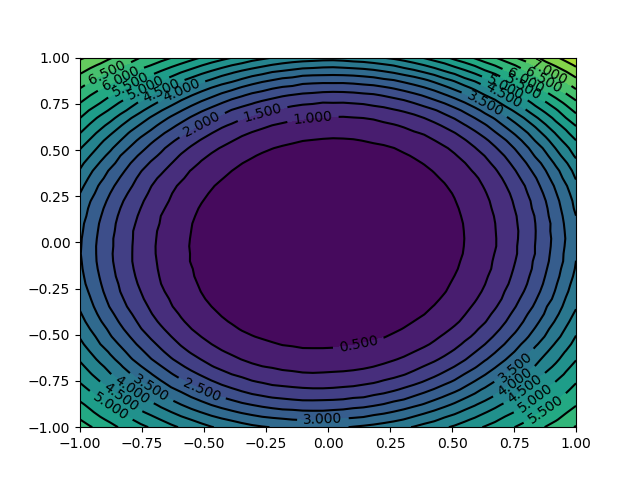}
			\caption{$128$; $87.78\%$}
			\label{fig:contour_plot1}
		\end{subfigure}
		\begin{subfigure}[b]{0.23\textwidth}
			\includegraphics[width=\textwidth]{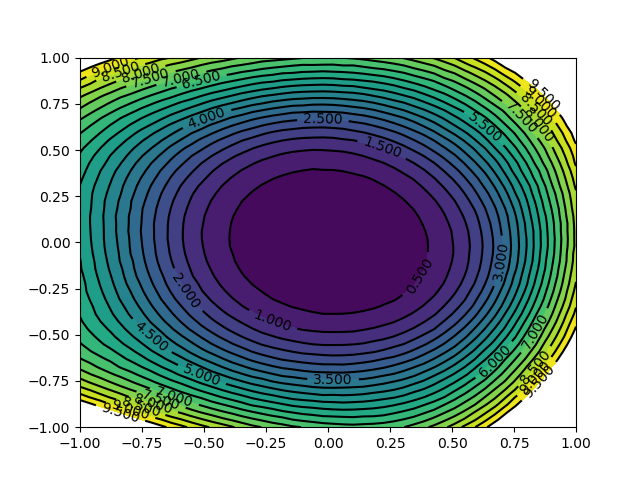}
			\caption{$2048$; $86.3\%$}
			\label{fig:contour_plot2}
		\end{subfigure}
		\caption{"128; $87.78\%$" refers to the model trained with $128$ batch size and achieved $87.78\%$ test accuracy; "2048; $86.3\%$" refers to the model trained with $2048$ batch size and achieved $86.3\%$ test accuracy; Figure (a)(b) shows the 3D plots for $f(t_1,t_2)$ with $(t_1, t_2)\in [-1,1]\times [-1,1] $ and Figure (c)(d) shows the contour maps for $f(t_1,t_2)$ with $(t_1, t_2)\in [-1,1]\times [-1,1] $.}
		\label{fig:gflatness}
	\end{figure*}
	Since we want to verify whether the generalization ability is related to PSI-flatness, 
	the first step is to obtain two minima with different generalization performance. It's widely accepted that SGD with small batch size produces minima generalize better than SGD with large batch size \cite{loss-surface2,SharpMinima}. Thus we train two ReLU NNs with the same structure via different batch sizes of SGD.
	
	The model we choosen is a stacked deep CNN-PlainNet-18 with ReLU activations \cite{he2016deep}. We use SGD to train the model on CIFAR-10 dataset \cite{krizhevsky2009learning}. The training strategy is same as the experiments reported in Section 5 of \cite{GSGD} except the batch size. 
	The batch sizes are chosen as $128$ and $2048$ respectively. 
	Both of the two experiments can make the training losses smaller than $10^{-3}$, which are close to zero. The test accuracy for the model trained with batch size $128$ is $87.78\%$; while the test accuracy for the model trained with batch size $2048$ is $86.3\%$.

	\par
	In order to visualize the loss landscapes represented by basis path values, we independently sample two random vectors $\xi$ and $\eta$ from Gaussian distribution $0.05\cdot\mathcal{N}(0, I)$. 
	Then for given $(t_1,t_2)$, we compute the value of $f(t_1,t_2)$.
	
	A simple method to obtain the value of $f(t_1,t_2)$ is to leverage the feed forward calculation. Since the explicit expression of loss based on values of basis path is computational cost, we project the perturbed values of basis paths back to weights in order to calculate $f(t_{1},t_{2})$. Here we leverage the skeleton method in Meng et al.\cite{GSGD} to design the projection. 
	
	In general, under the framework of skeleton method, there exists a bijection $H(w): \mathcal{W}\rightarrow \mathcal{V}$ between weights and values of basis paths when the values of some weights are fixed. So if we perturb the basis path values as $v = v + \varepsilon$, the resulting projection in weights can be obtained as
	\begin{equation}
	w + \delta= H^{-1}(v +\varepsilon),
	\end{equation}
	where $\delta$ is the corresponding perturbation in weight space.

	
	For a basis path that contains no non-skeleton weight, we project its perturbed value $v_i+\epsilon_i$ to the un-fixed weight. Because we have $v_i=w_i$, then by solving $\delta_i$ from $v_i+\epsilon_i=(w_i+\delta_i)\cdot \frac{v_i}{w_i}$, we have 
	\begin{equation}
	w_i+\delta_i=\frac{w_i\cdot(v_{i}+\varepsilon_{i})}{v_i},
	\label{eq:biases1}
	\end{equation}
	Similarly, we project the value of 	a basis path that contains one non-skeleton weight to the non-skeleton weight $w_j$. We have
	\begin{equation}
	w_j+\delta_j=\frac{w_j+\epsilon_j\cdot\frac{w_j}{v_j}}{1+\delta_{i(j)}/w_{i(j)}}, 
	\label{eq:biases2}
	\end{equation} where $w_{i(j)}$ denotes the skeleton weight in path $v_j$ that has been updated in Equation (\ref{eq:biases1})\footnote{According to the theory  in the work \cite{GSGD}, for the basis path that contains a non-skeleton weight, there will be only one skeleton weight that is not free variables with fixed value.} . 
	
	
	We can directly calculate the new weight $w_i+\delta_i$ and $w_j+\delta_j$ after the projection according to equation \eqref{eq:biases1} and equation \eqref{eq:biases2}. The value of $f(t_1,t_2)$ can be easily obtained by forward propagation for every $(t_1,t_2)$. 
	
	For the two trained models, we show two kinds of figures: the 3D plots of $f(t_1,t_2)$ (shown in \ref{fig:contour_plot1}) and the contour maps of $f(t_{1},t_{2})$ (shown in \ref{fig:contour_plot2}). All the pictures are plotted with same height interval. 
	
	\par
	From the figures (\ref{fig:gflatness}), we obviously observe that the minimum with better test accuracy locates in a "flatter" region under PSI-flatness measurement. It reveals that PSI-flatness is indeed related to the generalization which is accord with our theoretically result. 

	\section{Conclusion}
	This paper focuses on the ill-posed problem of flatness proposed in Dinh et al.\cite{SharpMinimacan}. We give the geometrical concept flatness a new definition and name the measure as PSI-flatness. We prove that PSI-flatness is a positively scale-invariant measure for ReLU NN and thus it is more suitable to study generalization. We also analyze the relationship between PSI-flatness and generalization. More specifically, we quantitatively build a connection between the PSI-flatness and generalization error under PAC Bayes theory. Then we give an upper bound to PSI-flatness which constructs an angle to study generalization from basis path perspective. Finally, we visualize the loss surface in PSI space and show that PSI-flat minima indeed generalize better, while this fails to appear for some cases in weight space. In the future, we will study the influence of other reparametrizations of NN models on the geometrical measures. 
	


	\bibliographystyle{abbrv}
	\bibliography{reference} 
	
	
	\appendix
	\section{Proof of Theorem 5.1}\vbox{}
	First, we proof the Lemma \ref{lem:repr} in the main paper. These proof is under the framework of skeleton method. Without loss of generality, we set the fixed skeleton weight value as 1.
	\par
	\noindent\textbf{Lemma 5.1 }
		\emph{For a neural network with $L$-layers and $m-H$ basis paths, the non-basis path value can be represented as:
		\begin{equation}
		\frac{v_{1}^{a_{1}}\cdots v_{m_{1}}^{a_{m_{1}}}}{v_{m_{1}+1}^{a_{m_{1}+1}}\cdots v_{m-H}^{a_{m-H}}},
		\end{equation}
		where $v_{1},\cdots,v_{m-H}$ are path value of basis path. $a_{1},\cdots,a_{m-H}$ in equation \ref{eq:non-basispath} are non-negative integers satisfy $a_{1}+\cdots+a_{m_{1}}=L+1$ and $a_{m_{1}+1}+\cdots+a_{m-H}=L$.}  
	\par 
	\vbox{}
	\noindent\textbf{Proof of Lemma \ref{lem:repr}:}
		By the construction of skeleton weight, we see that weight parameter not in the first layer can be represent as the ratio between two basis path values, which respectively are one basis path only containing skeleton weight and a basis path containing one non-skeleton weight. Mathematically speaking, weight parameters not in the first layer equal to $\frac{p_{i_{1}}}{p_{i_{2}}}$, where $p_{i_{\cdot}}$ are basis path values. The weight in the first layer is directly represented by one basis path value. Hence, the definition of path value $\prod_{l=1}^{L}w_{l}(i_{l-1},i_{l})$ implies the conclusion.$\hfill\blacksquare$
		\par
	Now we give the complete proof of the Theorem \ref{thm:bound}. According to skeleton method, if $v = v + \varepsilon$, then we note $w + \delta= H^{-1}(v +\varepsilon)$. We have
	\begin{equation}
	\begin{aligned}
	w_i+\delta_i=v_{i}+\varepsilon_{i}=w_{i}+\varepsilon_{i},\\
	w_j+\delta_j=\frac{w_j+\epsilon_j/w_{i(j)}}{1+\delta_{i(j)}/w_{i(j)}}.
	\end{aligned}
	\end{equation} 
	Here $w_{i}$ is the skeleton weight in the first layer, while $w_{j}$ is the non-skeleton weight.$w_{i(j)}$ is the skeleton in the first layer locate in the same basis path $p_{j}$ with non-skeleton weight $w_{j}$ and $\delta_{i(j)}$ is the perturbation of $w_{i(j)}$ in weight space.
	\par
	\vbox{}
	\noindent\textbf{Theorem 5.1 }
		\emph{Under Assumption \ref{A1} and \ref{A2}, for a $L$-hidden layer ReLU neural network with $d_{0}$ dimensional input and width $d_1$ for each hidden layer, we have
		\begin{equation}
		\begin{aligned}
		\mathop{\max}_{v^{\prime}\in B_{2}(v,\varepsilon)}\left(\hat{l}(v^{\prime})-\hat{l}(v)\right)\leq
		\quad2C\cdot C_{L}\varepsilon\cdot d_0\cdot d_{1}^{L-1}\cdot d_{L}\left(1+2^{\frac{3L+4}{2}}(L-1)d_{1}d_L\psi(v)\right),
		\end{aligned}
		\end{equation}
		where $\psi(v)=\sup_{s\in[L]}\varepsilon^{s-1}\frac{\sup_{i\in[m-H]}|v_{i}|^{2L+1-s}}{\inf_{i\in[m-H]}|v_{i}|^{2L}}$ \footnote{$\sup_{s\in[L]}$ denotes the maximum among $s=1$, $s=2,\cdots,s=L$} and $v_{i}$ is the $i$-th basis path value, when $\varepsilon$ is smaller than a constant decided by path value and input data.}\par
	\vbox{}
	\noindent\textbf{Proof of Theorem \ref{thm:bound}:}. 
	Without loss of generality, we assume the upper bound $C$ in Assumption \ref{A1} is 1, the proof is exactly unchanged. Suppose $\mathop{\max}_{v^{\prime}\in B_{2}(v,\varepsilon)}\left(\hat{l}(v^{\prime})-\hat{l}(v)\right)$ is attained at $v^{\prime}$ and set $v^{\prime}-v=\delta$ with $\|\delta\|\leq \varepsilon$. The symbols follow the main text part. For any input data $x$, we first give the appropriate scale of $\varepsilon$ to keep $D^{\prime}_{i}=D_{i}$. We notice that 
	\begin{equation}
	\begin{aligned}
	\|D_{1}^{\prime}-D_{1}\|^{2}&=\sum_{i=1}^{d_1}\left|\mathbb{I}\left(\sum_{j=1}^{d}(v_{j}^{i}+\delta_{j}^{i})\cdot x_{j}\geq 0\right)-\mathbb{I}\left(\sum_{j=1}^{d}v_{j}^{i}\cdot x_{j}\geq 0\right)\right|\\
	&= \sum_{i=1}^{d_1}\mathbb{I}\left(\sum_{j=1}^{d}(v_{j}^{i}+\delta_{j}^{i})\cdot x_{j}\cdot\sum_{j=1}^{d}v_{j}^{i}\cdot x_{j}\leq 0\right),
	\end{aligned}
	\label{eq:signal}
	\end{equation}
	where $v_{j}^{i}$ is the value of basis path connecting the $i$-th hidden neuron in layer two and the $j$-th input neuron in the first layer, and $\delta_{j}^{i}$ is the perturbation to corresponding path value. We notice that
	\begin{equation}
	\begin{aligned}
	\sum_{i=1}^{d_1}\mathbb{I}&\left[\left(\sum_{j=1}^{d}(v_{j}^{i}+\delta_{j}^{i})\cdot x_{j}\right)\cdot\left(\sum_{j=1}^{d}v_{j}^{i}\cdot x_{j}\right)\leq 0\right] \leq \sum_{i=1}^{d_1}\mathbb{I}\left(\left|\sum_{j=1}^{d}\delta_{j}^{i}\cdot x_{j}\right|\geq \left|\sum_{j=1}^{d}v_{j}^{i}\cdot x_{j}\right|\right)\\
	&\leq \sum_{i=1}^{d_1}\mathbb{I}\left(\left|\sum_{j=1}^{d}v_{j}^{i}\cdot x_{j}\right|\leq \|x\|\cdot\|\delta^{i}\|\right)\leq\sum_{i=1}^{d_1}\mathbb{I}\left(\left|\sum_{j=1}^{d}v_{j}^{i}\cdot x_{j}\right|\leq \|\delta\|\right),
	\end{aligned}
	\end{equation}
	where we assume that the input to each neuron is not exactly zero. Choosing $\|\delta\|\leq\inf_{x, i}\left|\sum_{j=1}^{d}v_{j}^{i}\cdot x_{j}\right|$, then we have $\|D_{1}-D_{1}^{\prime}\|=0$. 
	\par
	To make sure $\|D_{2}^{\prime}-D_{2}\|=0$, we have
	\begin{equation}
	\|D_{2}^{\prime}-D_{2}\| = \sum_{i=1}^{d_1}\left|\mathbb{I}\left(\hat{y}_{1}+\sum_{j=2}^{d_1}\frac{(v_{j}^{i}+\delta_{j}^{i})\hat{y}_{j}}{v_{j}^{i}(i,j)+\delta_{j}^{i}(i,j)}\geq 0\right)-\mathbb{I}\left(y_{1}+\sum_{j=2}^{d_1}\frac{v_{j}^{i}}{v_{j}^{i}(i,j)}y_{j}\geq 0\right)\right|,
	\label{eq:D2}
	\end{equation}
	where $\hat{y}_{j}, \hat{y}_{j}$ are respectively outputs and outputs after perturbed of the $j$-th hidden neurons in the second layer. And $v_{j}^{i}$ in equation \eqref{eq:D2} is the path value of basis path connecting the $i$-th neuron in layer 3 and the the $j$-th neuron in layer 2, $\delta_{(\cdot)}^{(\cdot)}$ are corresponding perturbation. And finally, the $v_{j}^{i}(i,j)$ and $\delta_{j}^{i}(i,j)$ in equation \eqref{eq:D2} is the path value of basis path containing skeleton weight only which cross path $p_{i}^{j}$ and its perturbation. The result is derived by the skeleton method. Then we have
	\begin{equation}
	\begin{aligned}
	\|D_{2}^{\prime}-D_{2}\|&=\sum_{i=1}^{d_2}\mathbb{I}\left[\left(\hat{y}_{1}+\sum_{j=2}^{d_1}\frac{(v_{j}^{i}+\delta_{j}^{i})\hat{y}_{j}}{v_{j}^{i}(i,j)+\delta_{j}^{i}(i,j)}\right)\left(y_{1}+\sum_{j=2}^{d_1}\frac{v_{j}^{i}}{v_{j}^{i}(i,j)}y_{j}\right)\leq 0\right]\\
	&\leq \sum_{i=1}^{d_2}\mathbb{I}\left(\left|\hat{y}_{1}+\sum_{j=2}^{d_1}\frac{v_{j}^{i}\hat{y}_{j}}{v_{j}^{i}(i,j)+\delta_{j}^{i}(i,j)}\right|\leq\left|\sum_{j=2}^{d_1}\frac{\delta_{j}^{i}\hat{y}_{j}}{v_{j}^{i}(i,j)+\delta_{j}^{i}(i,j)}\right|\right)\\
	&\leq \sum_{i=1}^{d_2}\mathbb{I}\left(\left|\hat{y}_{1}+\sum_{j=2}^{d_1}\frac{v_{j}^{i}\hat{y}_{j}}{v_{j}^{i}(i,j)+\delta_{j}^{i}(i,j)}\right|\leq \left(\sum_{j=2}^{d_1}(\delta_{j}^{i})^{2}\right)^{\frac{1}{2}}\left(\frac{(\hat{y}_{j}^{i})^{2}}{\sum_{j=2}^{d_1}(v_{j}^{i}(i,j)+\delta_{j}^{i}(i,j))^{2}}\right)^{\frac{1}{2}}\right)\\
	&\leq \sum_{i=1}^{d_2}\mathbb{I}\left(\frac{\left|\hat{y}_{1}+\sum_{j=2}^{d_1}\frac{v_{j}^{i}\hat{y}_{j}}{v_{j}^{i}(i,j)+\delta_{j}^{i}(i,j)}\right|}{\left[\frac{(\hat{y}^{i}_{j})^{2}}{\sum_{j=2}^{d_1}(v_{j}^{i}(i,j)+\delta_{j}^{i}(i,j))^{2}}\right]^{\frac{1}{2}}}\leq\|\delta\|\right).
	\end{aligned}
	\label{eq:D2_out}
	\end{equation}
	Now, we analysis the last term in equation \eqref{eq:D2_out}. If we choosing $\delta$ to make the activation status unchanged in the second layer, then we have
	\begin{equation}
	|\hat{y}_{i}-y_{i}|=\left|\sum_{j=1}^{d_1}\delta_{j}^{i}x_{j}\right|\leq\|\delta\|\|x\|=\|\delta\|.
	\label{eq:y_bound}
	\end{equation}
	We see the
	\begin{equation}
	\begin{aligned}
	\frac{\left|\hat{y}_{1}+\sum_{j=2}^{d_1}\frac{v_{j}^{i}\hat{y}_{j}}{v_{j}^{i}(i,j)+\delta_{j}^{i}(i,j)}\right|}{[\sum_{j=2}^{d_1}(v_{j}^{i}(i,j)+\delta_{j}^{i}(i,j))^{2}]^{\frac{1}{2}}}&\leq \frac{\left|y_{1}+\sum_{j=2}^{d_1}\frac{v_{j}^{i}y_{j}}{v_{j}^{i}(i,j)+\delta_{j}^{i}(i,j)}\right|-\left|\hat{y}_{1}-y_{1}+\sum_{j=2}^{d_1}\frac{v_{j}^{i}(\hat{y}_{j}-y_{j})}{v_{j}^{i}(i,j)+\delta_{j}^{i}(i,j)}\right|}{\left[2\sum_{j=2}^{d_1}\left(\frac{\hat{y}_{j}^{i}-y_{j}^{i}}{v_{j}^{i}(i,j)+\delta_{j}^{i}(i,j)}\right)^{2}+2\sum_{j=2}^{d_1}\left(\frac{y_{j}^{i}}{v_{j}^{i}(i,j)+\delta_{j}^{i}(i,j)}\right)^{2}\right]^{\frac{1}{2}}}\\
	&\leq \frac{\left|y_{1}+\sum_{j=2}^{d_1}\frac{v_{j}^{i}y_{j}}{v_{j}^{i}(i,j)+\delta_{j}^{i}(i,j)}\right|-\|\delta\|\left(1+\left|\sum_{j=2}^{d_1}\frac{v_{j}^{i}}{v_{j}^{i}(i,j)+\delta_{j}^{i}(i,j)}\right|\right)}{\left[2\|\delta\|\sum_{j=2}^{d_1}\left(\frac{1}{v_{j}^{i}(i,j)+\delta_{j}^{i}(i,j)}\right)^{2}+2\sum_{j=2}^{d_1}\left(\frac{y_{j}^{i}}{v_{j}^{i}(i,j)+\delta_{j}^{i}(i,j)}\right)^{2}\right]^{\frac{1}{2}}}.
	\end{aligned}
	\label{eq:delta}
	\end{equation}
	Choosing $\|\delta\|$ smaller than the last term in the above equation will make $\|D_{2}^{\prime}-D_{2}\|=0$. For any $\eta>0$, there exist a $\delta_{0}(\eta)>0$ satisfy 
	\begin{equation}
	\left|\frac{\left|y_{1}+\sum_{j=2}^{d_1}\frac{v_{j}^{i}y_{j}}{v_{j}^{i}(i,j)}\right|}{\left[2\sum_{j=2}^{d_1}\left(\frac{y_{j}^{i}}{v_{j}^{i}(i,j)}\right)^{2}\right]^{\frac{1}{2}}} - \frac{\left|y_{1}+\sum_{j=2}^{d_1}\frac{v_{j}^{i}y_{j}}{v_{j}^{i}(i,j)+\delta_{j}^{i}(i,j)}\right|-\|\delta\|\left(1+\left|\sum_{j=2}^{d_1}\frac{v_{j}^{i}}{v_{j}^{i}(i,j)+\delta_{j}^{i}(i,j)}\right|\right)}{\left[2\|\delta\|\sum_{j=2}^{d_1}\left(\frac{1}{v_{j}^{i}(i,j)+\delta_{j}^{i}(i,j)}\right)^{2}+2\sum_{j=2}^{d_1}\left(\frac{y_{j}^{i}}{v_{j}^{i}(i,j)+\delta_{j}^{i}(i,j)}\right)^{2}\right]^{\frac{1}{2}}}\right|< \eta,
	\end{equation}
	when $\|\delta\|\leq\delta_{0}(\eta)$. Choosing $\eta<\inf_{y,i}\frac{\left|y_{1}+\sum_{j=2}^{d_1}\frac{v_{j}^{i}y_{j}}{v_{j}^{i}(i,j)}\right|}{\left[2\sum_{j=2}^{d_1}\left(\frac{y_{j}^{i}}{v_{j}^{i}(i,j)}\right)^{2}\right]^{\frac{1}{2}}}$ and
	\begin{equation}
	\|\delta\|<\min\left(\inf_{y,i}\frac{\left|y_{1}+\sum_{j=2}^{d_1}\frac{v_{j}^{i}y_{j}}{v_{j}^{i}(i,j)}\right|}{\left[2\sum_{j=2}^{d_1}\left(\frac{y_{j}^{i}}{v_{j}^{i}(i,j)}\right)^{2}\right]^{\frac{1}{2}}}-\eta, \delta_{0}(\eta), \inf_{x, i}\left|\sum_{j=1}^{d}v_{j}^{i}\cdot x_{j}\right|\right)
	\end{equation}
	will keep the activation status unchanged. The assumption $A_{2}$ ensure the existence of $\|\delta\|\cdot C$. The proof of $\|D_{i}^{\prime}-D_{i}\|=0$ can be achieved by the proof of $\|D_{2}^{\prime}-D_{2}\|=0$ without any change. 
	\par
	Now, we make sure the activation status unchanged. Then according to the Lipschitz continuous of $L(v, x)$ (Assumption $A_{2}$), we have
	\begin{equation}
	\begin{aligned}
	|l(v^{\prime},x)-(v,x)|&\leq C_{L}\|f_{v^{\prime}}(x) - f_v(x)\|\\
	&\leq C_{L}\left|\sum_{k=1}^{K}\left(\sum_{i}\sum_{p\in\mathcal{P}_{k,i}}(v(p)-v(p^{\prime}))\cdot a(p,x) \cdot x_{i}\right)^{2}\right|^{\frac{1}{2}}\\
	&\leq C_{L}\left(n_{p}^{2}\sum_{k=1}^{K}\|x\|^{2}\left(\sum_{i}\sum_{p\in\mathcal{P}_{k,i}}\delta_{p}^{2}\right)\right)^{\frac{1}{2}},
	\end{aligned}
	\label{eq:out}
	\end{equation}
	Where $\delta_{p}$ is the perturbation on path value $p$, $n_{p}$ is the number of path, $\mathcal{P}_{k}$ is path connecting the $k$-th output node. Since
	\begin{equation}
	\sum_{k=1}^{K}\sum_{i}\sum_{p\in\mathcal{P}_{k,i}}\delta_{p}^{2} = \sum_{k=1}^{K}\sum_{i}\sum_{p\in\mathcal{P}_{k,i}\bigcap\mathcal{S}_{k,i}}\delta_{p}^{2} + \sum_{k=1}^{K}\sum_{i}\sum_{p\in\mathcal{P}^{k,i}\bigcap\mathcal{S}_{k,i}^{c}}\delta_{p}^{2}\leq \|\delta\|^{2} + \sum_{k=1}^{K}\sum_{i}\sum_{p\in\mathcal{P}^{k,i}\bigcap\mathcal{S}_{k,i}^{c}}\delta_{p}^{2},
	\label{eq:biases} 
	\end{equation}
	where $\mathcal{S}^{k,i}$ is the basis path connecting the $k$-th output node and the $i$-th input. By equation \eqref{eq:non-basispath}, for any $p\in\mathcal{P}^{k,i}\bigcap\mathcal{S}_{k,i}^{c}$, we have
	\begin{equation}
	\begin{aligned}
	|\delta_{p}| &= \left|\frac{\prod_{i=1}^{L+1}(v_{p_{i}}+\delta_{p_{i}})}{\prod_{i=L+2}^{2L+1}(v_{p_{i}}+\delta_{p_{i}})} - \frac{\prod_{i=1}^{L+1}v_{p_{i}}}{\prod_{i=L+2}^{2L+1}v_{p_{i}}}\right| \\
	&= \left|\frac{\prod_{i=1}^{L+1}(v_{p_{i}}+\delta_{p_{i}})\prod_{i=L+2}^{2L+1}v_{p_{i}} - \prod_{i=1}^{L+1}v_{p_{i}}\prod_{i=L+2}^{2L+1}(v_{p_{i}}+\delta_{p_{i}})}{\prod_{i=L+2}^{2L+1}(v_{p_{i}}+\delta_{p_{i}})\prod_{i=L+2}^{2L+1}v_{p_{i}}}\right|
	\end{aligned}.
	\end{equation}
	The last term of the above equation can be bounded by the summation of $2^{L+1}+2^{L}$ items(one of them is zero). We get the result by split the formulation in numerator of the equation. Each of them has the form like
	\begin{equation}
	\left|\frac{v_{p_{1}}\cdots\delta_{p_{i}}\cdots v_{2L+1}}{\prod_{i=L+2}^{2L+1}(v_{p_{i}}+\delta_{p_{i}})\prod_{i=L+2}^{2L+1}v_{p_{i}}}\right|.
	\label{eq:frac}
	\end{equation}
	We bound \eqref{eq:frac} to illustrate our method, other terms can be achieved follow the same procedure.
	\begin{equation}
	\begin{aligned}
	\left|\frac{v_{p_{1}}\cdots\delta_{p_{i}}\cdots v_{2L+1}}{\prod_{i=L+2}^{2L+1}(v_{p_{i}}+\delta_{p_{i}})\prod_{i=L+2}^{2L+1}v_{p_{i}}}\right| &\leq \left|\frac{v_{p_{1}}\cdots\delta_{p_{i}}\cdots v_{2L+1}}{\prod_{i=L+2}^{2L+1}\sqrt{\frac{v_{p_{i}}^{2}+\delta_{p_{i}}^{2}}{2}}\prod_{i=L+2}^{2L+1}v_{p_{i}}}\right|\\
	&\leq \left|2^{\frac{L}{2}}\frac{v_{p_{1}}\cdots\delta_{p_{i}}\cdots v_{2L+1}}{\prod_{i=L+2}^{2L+1}v_{p_{i}}^{2}}\right|\\
	&\leq 2^{\frac{L}{2}}\|\delta\|\frac{\sup_{v}|v_{p}|^{2L}}{{\inf_{v}|v_{p}|^{2L}}.}
	\end{aligned}
	\end{equation}
	Following the procedure, we can actually get result that every equation like \eqref{eq:frac} owns an upper bound perform like $2^{\frac{L}{2}}\|\delta\|^{\alpha}\frac{\sup_{v}|v_{p}|^{2L+1-\alpha}}{\inf_{v}|v_{p}|^{2L}}$, where $\alpha=\{1,\cdots, L+1\}$. Hence, we conclude
	\begin{equation}
	|\delta_{p}|\leq 2^{\frac{L}{2}}(2^{L+1}+2^{L}-1)\sup_{\alpha=\{1,\cdots,L+1\}}\varepsilon^{\alpha}\frac{\sup_{v}|v_{p}|^{2L+1-\alpha}}{\inf_{v}|v_{p}|^{2L}}.
	\label{eq:non_basis_path}
	\end{equation}
	Noticing that $n_{p}=d_{0}\cdot d_{1}^{L-1}\cdot d_{L}$ and the number of non-basis path is $(L-1)\cdot d_{1}$. Substituting \eqref{eq:non_basis_path} into \eqref{eq:out} and combining \eqref{eq:biases}. By the definition of $l(v)=\frac{1}{n}\sum_{i=1}^{n}l(v,x_{i})$, and the inequality $\sqrt{\frac{a^{2}+b^{2}}{2}}\leq |a|+|b|$, we get the conclusion
	
\end{document}